\documentclass[conference, a4paper]{IEEEtran}
\usepackage{flushend}
\usepackage{paralist}
\usepackage[group-separator={,}]{siunitx}
\usepackage{cite}
\ifCLASSINFOpdf
  \usepackage[pdftex]{graphicx}
  \graphicspath{{images/}}
\else
  \usepackage[dvips]{graphicx}
  \graphicspath{{images/}}
\fi
\usepackage{amsmath}
\DeclareMathOperator*{\argmax}{arg\,max}

\usepackage{array}
\ifCLASSOPTIONcompsoc
 \usepackage[caption=false,font=normalsize,labelfont=sf,textfont=sf]{subfig}
\else
 \usepackage[caption=false,font=footnotesize]{subfig}
\fi
\usepackage{url}
\begin{document}
\title{\LARGE \bf Skin Disease Classification versus Skin Lesion Characterization: Achieving Robust Diagnosis using Multi-label Deep Neural Networks}

\author{\IEEEauthorblockN{Haofu Liao, Yuncheng Li, Jiebo Luo}
\IEEEauthorblockA{Department of Computer Science, University of Rochester\\
Rochester, New York 14627, USA\\
Email: \{hliao6, yli, jluo\}@cs.rochester.edu}
}


%Telephone: (800) 555--1212, Fax: (888) 555--1212}


\maketitle

\begin{abstract}
In this study, we investigate what a practically useful approach is in order to achieve robust
skin disease diagnosis. A direct approach is to target the ground truth diagnosis
labels, while an alternative approach instead focuses on determining skin lesion
characteristics that are more visually consistent and discernible. We argue that,
for computer aided skin disease diagnosis, it is both more realistic and more
useful that lesion type tags should be considered as the target of an automated
diagnosis system such that the system can first achieve a high accuracy in
describing skin lesions, and in turn facilitate disease diagnosis using lesion
characteristics in conjunction with other evidences. To further meet such an
objective, we employ convolutional neutral networks (CNNs) for both the 
disease-targeted and lesion-targeted classifications. We have collected a
large-scale and diverse dataset of $\num{75665}$ skin disease images from six
publicly available dermatology atlantes. Then we train and compare both
disease-targeted and lesion-targeted classifiers, respectively. For
disease-targeted classification, only $\mathbf{27.6\%}$ top-$\mathbf{1}$
accuracy and $\mathbf{57.9\%}$ top-$\mathbf{5}$ accuracy are achieved with a mean
average precision (mAP) of $\mathbf{0.42}$. In contrast, for lesion-targeted
classification, we can achieve a much higher MAP of $\mathbf{0.70}$.
\end{abstract}


% \fi
\IEEEpeerreviewmaketitle

\section{Introduction}

The diagnosis of skin diseases is challenging. To diagnose a skin disease,
a variety of visual clues may be used such as the individual lesional 
morphology, the body site distribution, color, scaling and arrangement of lesions.
When the individual elements are analyzed separately, the recognition process
can be quite complex \cite{cox2004diagnosis}. For example,
the well studied skin cancer, melanoma, has four major clinical
diagnosis methods: ABCD rules, pattern analysis, Menzies method and $7$-Point 
Checklist. To use these methods and achieve a satisfactory diagnostic accuracy, a high
level of expertise is required as the differentiation of skin lesions demands 
a great deal of experience and expertise \cite{whited1998does}.

Unlike the diagnosis by human experts, which depends essentially on subjective 
judgment and is not always reproducible, a computer aided diagnostic system is more
objective and reliable. Traditionally, one can use human-engineered feature
extraction algorithms in combination with a classifier to complete this
task. For some skin diseases, such as melanoma and basal cell carcinoma, this
solution is feasible as their features are regular and predictable. However,
when we extend the skin diseases to a broader range, where the features are so
complex that hand-crafted feature design becomes infeasible, the traditional
approach fails.

In recent years, deep convolutional neural networks (CNN) become very popular 
in feature learning and object classification. The use of high performance GPUs
makes it possible to train a network on a large-scale dataset so as to
yield a better performance. Many studies 
\cite{DBLP:journals/corr/SermanetEZMFL13,DBLP:journals/corr/IoffeS15,DBLP:journals/corr/SzegedyLJSRAEVR14,DBLP:journals/corr/SimonyanZ14a} 
from the ImageNet Large Scale Visual Recognition Challenge (ILSVRC) \cite{ILSVRC15} 
have shown that the state-of-art CNN architectures are able to surpass humans in many 
computer vision tasks. Therefore, we propose to construct a skin disease classifier
with CNNs.

\begin{figure}[t]
  \centering
  \includegraphics[scale=0.25]{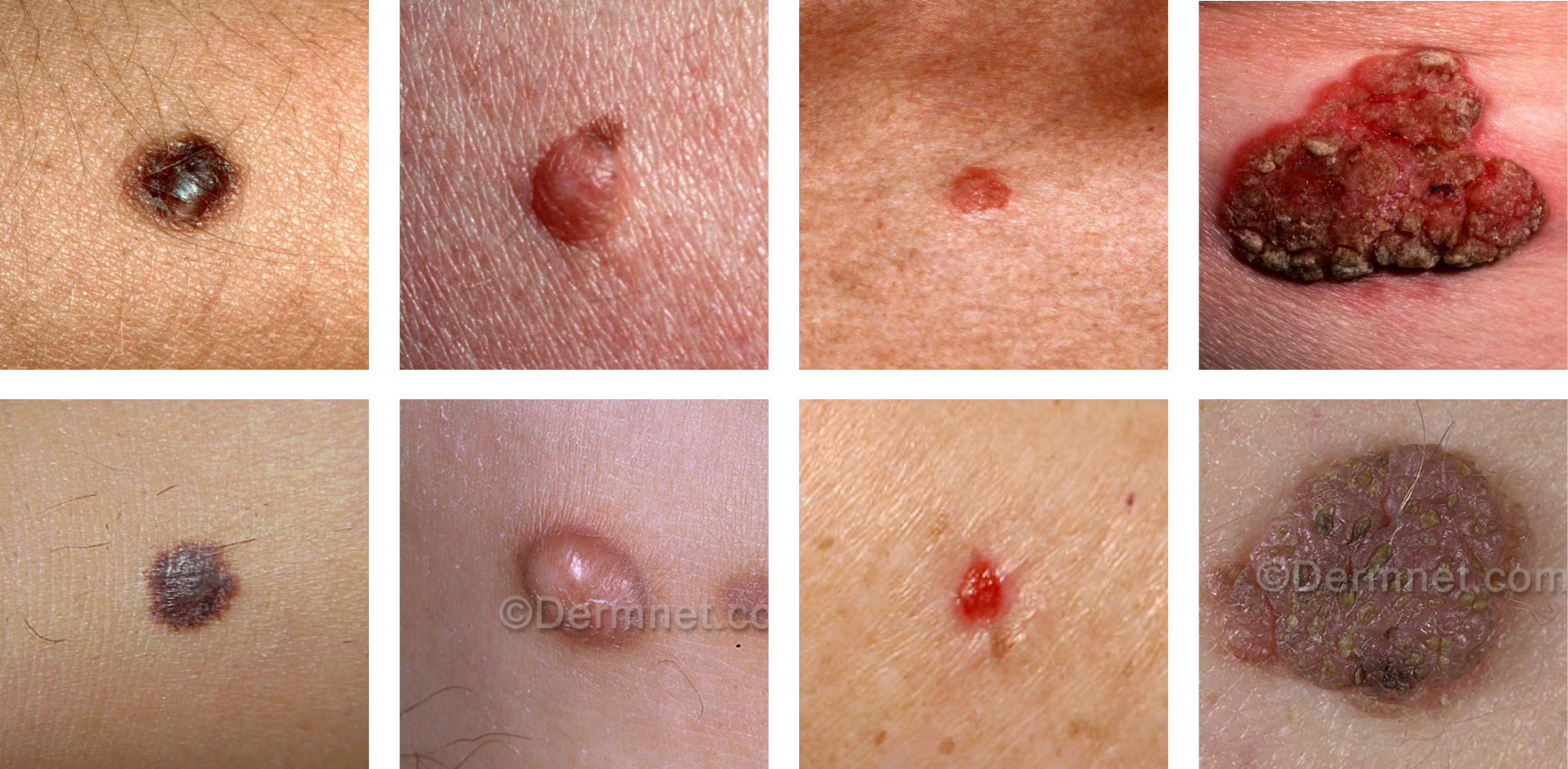}
  \caption{Some visually similar skin diseases. First row (left to right):
  malignant melanoma, dermatofibroma, basal cell carcinoma, seborrheic keratosis.
  Second row (left to right): compound nevus, intradermal nevus, benign keratosis,
  bowen's disease.}
  \label{fig: similar diseases}
  \vskip -0.1in
\end{figure}

However, training CNNs  directly using the diagnosis labels may not be viable.
\begin{inparaenum}[1)]
  \item For some diseases, their lesions are so similar that they can not be
  distinguished visually. Figure \ref{fig: similar diseases} shows the
  dermatology images of eight different skin diseases. We can see that
  the two diseases in each column have very similar visual appearances. Thus, it
  is very difficult to make a judgment between the two diseases with only the visual
  information.
  \item Many of the skin diseases are not so common that only a few images are available
  for training. Table \ref{tab: diagnosis statistics} shows the dataset statistics
  of the dermatology atlantes we used in this study. We can see that there are
  tens of hundreds of skin diseases. However, most of them contain very few
  images.
  \item Skin disease diagnosis is a complex procedure that often involves many other
  modalities, such as palpation, smell, temperature changes and microscopy
  examinations \cite{cox2004diagnosis}.
\end{inparaenum}

On the other hand, lesion characteristics, which inherently describe the visual aspects of
skin diseases, arguably should be considered as the ideal ground truth for training.
For example, the two images in the first column of Figure \ref{fig: similar diseases}
can both be labeled with hyperpigmented and nodular lesion tags. Compared with
using the sometimes ambiguous disease diagnosis labels for these two images, the use of the
lesion tags can give a more consistent and precise description of the
dermatology images.

In this paper, we investigate the performance of CNNs trained with disease and
lesion labels, respectively. We collected $\num{75665}$ skin disease images from six
different publicly available dermatology atlantes. We then train a multi-class
CNN for  disease-targeted classification and another multi-label CNN for 
lesion-targeted classification. Our experimental results show that the top-$1$
and top-$5$ accuracies for the disease-targeted classification are $27.6\%$ and
$57.9\%$ with a mean average precision (mAP) of $0.42$. While for the
lesion-targeted skin disease classification, a much higher mAP of $0.70$ is achieved.

\section{Related Work}

Much work has been proposed for computer aided skin disease classification.
However, most of them use human-engineered feature extraction algorithms and restrict
the problem to certain skin diseases, such as melanoma \cite{arroyo2014automated,xie2014dermoscopy,Fabbrocini:2014yq,Saez:2014qy,Barata:2013fj}.
Some other works \cite{cruz2014automatic,wang2014cascaded,arevalo2015unsupervised} 
use CNNs for unsupervised feature learning from histopathology images and only
focus on the detection of mitosis, an indicator of cancer. Recently,
\textit{Esteva et al.} \cite{estevadeep} proposed a disease-targeted skin
disease classification method using CNN. They used the dermatology images from the
Dermnet atlas, one of the six atlantes used in this study, and reported that their
CNN achieved $60.0\%$ top-1 accuracy and $80.3\%$ top-$3$ accuracy.
However, they performed the CNN training and testing on the same dataset without cross-validation which makes their results unpersuasive. A preliminary work \cite{liao2016deep} of this study has also discovered similar performances on skin disease classification.
% In our study, we train and test the CNN on independent datasets for
% the disease-targeted skin disease classification and the experimental results
% show a much lower accuracy.

\section{Datasets} \label{sec: datasets}

We collect dermatology photos from the following dermatology atlas websites:
\begin{itemize}
  \item \textbf{AtlasDerm} (www.atlasdermatologico.com.br)
  \item \textbf{Danderm} (www.danderm-pdv.is.kkh.dk)
  \item \textbf{Derma} (www.derma.pw)
  \item \textbf{DermIS} (www.dermis.net)
  \item \textbf{Dermnet} (www.dermnet.com)
  \item \textbf{DermQuest} (www.dermquest.com)
\end{itemize}
These atlantes are maintained by professional dermatology resource providers. 
They are used by dermatologists for training and teaching purpose. All of the
dermatology atlantes have diagnosis labels for their images. For each dermatology
image only one disease diagnosis label is assigned. We use these diagnosis labels
as the ground truth to train the disease-targeted skin disease classifier. 

However, each of the atlas maintains its own skin disease taxonomy and naming
convention for the diagnosis labels. It means different atlas may have different
labels for the same diagnosis and some diagnosis may have several variations.
To address this problem, we adapt the skin disease taxonomy used by the 
DermQuest atlas and merge the diagnosis labels from other atlantes into it. We
choose the DermQuest atlas because of the completeness and professionalism of
its dermatology resources. In most of the cases, the labels for the same diagnoses
may have similar naming conventions. Therefore, we merge them by looking at the
word or string similarity of two diagnosis labels. We use the string pattern matching
algorithm described in \cite{ratcliff1988pattern}, where the similarity ratio
is 
\begin{equation}
S = \frac{2 * M}{T}.
\end{equation}
Here, $M$ is the number of matches and $T$ is the total number of characters in
both strings. The statistics of the merged atlantes is given
in Table \ref{tab: diagnosis statistics}. Note that the total number of diagnoses
in our dataset is $\num{2113}$ which is significant higher than any of the atlas.
This is because we use a conservative merging strategy such that we merge two
diagnosis labels only when their string similarity is very high ($S > 0.8$).
Thus, we can make sure no two diagnosis labels are incorrectly merged. For those
redundant diagnosis labels, they only contain a few dermatology images. We can
discard them by choosing a threshold that filters out small diagnosis labels.

For the disease-targeted skin disease classification, we choose the AtlasDerm,
Danderm, Derma, DermIS, and Dermnet datasets as the training set and the
DermQuest dataset as the test set. Due to the inconsistency of the taxonomy and
naming convention between the atlantes, most of the diagnosis labels have only a few
images. As our goal is to investigate the feasibility of using CNNs for
disease-targeted skin disease classification, we remove these noisy diagnosis
labels and only keep those labels that have more than $300$ images. As a result
of the label refinement and cleaning, we have $\num{18096}$ images in the training set and
$\num{14739}$ images in the test set. The total number of diagnosis labels is $38$.

% \begin{table}
%   \renewcommand{\arraystretch}{1.3}
%   \caption{Major Lesion Tags}
%   \centering
%   \begin{tabular}{l|l}
%     \hline
%     \multicolumn{2}{c}{Major Lesion Tags from DermQuest} \\
%     \hline
%     0. Atrophy &             12. Nail Hyperpigmentation \\
%     1. Comedones &           13. Nodule \\ 
%     2. Crust &               14. Onycholysis \\ 
%     3. Edema &               15. Oozing \\
%     4. Erythemato-Squamous & 16. Pearly \\
%     5. Erythematous &        17. Scales \\
%     6. Excoriation &         18. Sclerosis \\     
%     7. Follicular &          19. Telangiectasis \\
%     8. Hyperkeratosis &      20. Tumour \\        
%     9. Hyperpigmented &      21. Ulceration \\    
%     10. Hypopigmented &      22. Vesicle \\       
%     11. Nail Dystrophy & 
%   \end{tabular}
%   \label{tab: lesion tags}
% \end{table}

For the skin lesions, only the DermQuest dataset contains the lesion tags.
Unlike the diagnosis, which is unique for each image, multiple lesion
tags may be associated with a dermatology image. There are a total of $134$
lesion tags for the $\num{22082}$ dermatology images from DermQuest. However, most
lesion tags have only a few images and some of the lesion tags are duplicated.
After merging and removing infrequent lesion tags, we retain $23$ lesion tags.

\begin{table}
  \renewcommand{\arraystretch}{1.3}
  \caption{Dataset statistics}
  \centering
  \begin{tabular}{l|c|c}
    \textbf{Atlas} & \textbf{\# of Images} & \textbf{\# of Diagnoses} \\
    \hline\hline
    AtlasDerm      &            \num{8766} &                   478   \\
    Danderm        &            \num{1869} &                    97   \\
    Derma          &           \num{13189} &                  1195   \\
    DermIS         &            \num{6588} &                   651   \\
    Dermnet        &           \num{21861} &                   488   \\
    DermQuest      &           \num{22082} &                   657   \\
    \hline\hline
    \textbf{Total} &  \textbf{\num{75665}} &         \textbf{2113} \\
  \end{tabular}
  \label{tab: diagnosis statistics}
\end{table}

Since only the DermQuest dataset has the lesion tags, we use images from the
DermQuest dataset to perform training and testing. The total number of
dermatology images that have lesion tags is $\num{14799}$. As the training
and test sets are sampled from the same dataset, to avoid overfitting, we use $5$-fold
cross-validation in our experiment. We first split our dataset into $5$ evenly
sized, non-overlapping ``folds''. Next, we rotate each fold as the test set and use
the remaining folds as the training set.

\section{Methodology} \label{sec: methodology}

We use CNNs for both the disease-targeted and lesion-targeted skin disease classifications.
For the disease-targeted classification, a multi-class
image classifier is trained and for the lesion-targeted classification,
we train a multi-label image classifier.

Our CNN architecture is based on the AlexNet \cite{NIPS2012_4824} and we modify 
it according to our needs. The AlexNet architecture was one of the early
wining entry of the ILSVRC challenges which is considered sufficient for this study. Readers
may refer to the latest winning entry (MSRA \cite{He2015} as of ILSVRC 2015)
for better performance. Implementation details of training and testing the CNNs
are given in the following sections.

\subsection{Disease-Targeted Skin Disease Classification} \label{sec: diagnosis MC}

For the disease-targeted skin disease classification, each dermatology image
is associated with only one disease diagnosis. Hence, we train a multi-class classifier
using CNN. We fine-tune the CNN with the BVLC AlexNet model \cite{jia2014caffe}
which is pre-trained from the ImageNet dataset \cite{ILSVRC15}. Since the
number of classes we are predicting is different with the ImageNet images, we
replace the last fully-connected layer ($1000$ dimension) with a new 
fully-connected layer where the number of outputs is set to the number of skin
diagnoses in our dataset.  We also increase the learning rate of the weights and
bias of this layer as the parameters of the newly added layer is randomly
initialized. For the loss function, we use the softmax function
\cite[Chapter~3]{nielsen2015neural} and connect a new softmax layer to the newly
added fully-connected layer. Formally put, let $z_j^L$ be the the weighted input
of the $j$th neuron of the softmax layer, where $L$ is the total number of the
layers in the CNN (For AlexNet, $L=9$). Thus, the $j$th activation of the
softmax layer is
\begin{equation}
a_j^L = \frac{e^{z_j^L}}{\sum_ke^{z_k^L}}
\end{equation}
And the corresponding softmax loss is
\begin{equation}
E = -\frac{1}{N} \sum_{n=1}^N\log(a_{y^n}^L)
\end{equation}
where $N$ is the number of images in a mini-batch, $y^n$ is the ground truth
of the $n$th image and $a_{y^n}^L$ is the $y^n$th activation of the softmax layer.
In the test phase, we choose the label $j$ that yields the largest activation
$a_j^L$ as the prediction, i.e.
\begin{equation}
\widehat{y} = \argmax_j{a_j^L}.
\end{equation}

\subsection{Lesion-Targeted Skin Disease Classification} \label{sec: lesion MLC}

As we mentioned early, multiple lesion tags may be associated with a
dermatology image. Therefore, to classify skin lesions we need to train a
multi-label CNN. Similar to disease-targeted skin disease classification,
we fine-tune the multi-label CNN with the BVLC AlexNet model. To train a
multi-label CNN, two data layers are required. One data layer loads the dermatology
images and the other data layer loads the corresponding lesion tags. Given
an image $\mathbf{X}_n$ from the first data layer, its corresponding lesion tags
from the second data layer are represented as a binary vector
$\mathbf{Y}_n=[y_1^n, y_2^n, \dots, y_Q^n]^T$
where $Q$ is the number of lesions in our data set and $y_j^n, j \in \{1, 2, \dots, Q\}$
is given as
\begin{equation}
y_j^n = \begin{cases}
  1, & \text{if the $j$th label is associated with $\mathbf{X}_n$,} \\
  0, & \text{otherwise.}
\end{cases}
\end{equation}
We replace the last fully-connected layer of the AlexNet with a new
fully-connected layer to accommodate the lesion tag vector. The learning rate of
the parameters of this layer is also increased so that the CNN can learn
features of the dermatology images instead of those images from ImageNet. For
the multi-label CNN, we use the sigmoid cross-entropy \cite[Chapter~3]{nielsen2015neural}
as the loss function and replace the softmax layer with a sigmoid cross-entropy
layer. Let the $z_j^L$ be the weighted input denoted in Section \ref{sec: lesion MLC},
then the $j$th activation of the sigmoid cross-entropy layer can be written as
\begin{equation}
a_j^L = \sigma(z_j^L) = \frac{1}{1 + e^{-z_j^L}}.
\end{equation}
And the corresponding cross-entropy loss is
\begin{equation}
E = - \frac{1}{N}\sum_{n=1}^N\sum_{j=1}^Qy_j^n\log{a_j^L} + (1-y_j^n)\log{(1 - a_j^L)}.
\end{equation}
For a given image $\mathbf{X}$, the output of the multi-label CNN is a
confidence vector $\mathbf{C}=[a_1^L, a_2^L, \dots, a_Q^L]^T$. Here,
$a_j^L$ is the $j$th activation of the sigmoid cross-entropy layer. It denotes
the confidence of $\mathbf{X}$ being related to the lesion tag $j$. In the
test phase, we use a threshold function $t(\mathbf{X})$ to determine the
lesion tags of the input image $\mathbf{X}$, i.e.
$\widehat{\mathbf{Y}} = [\widehat{y}_1, \widehat{y}_2, \dots, \widehat{y}_Q]^T$
where
\begin{equation}
\widehat{y}_j = \begin{cases}
  1, & a_j^L > t(\mathbf{X}), \\
  0, & \text{otherwise,}
\end{cases}
j \in \{1, 2, \dots, Q\}.
\end{equation}
For the choice of the threshold function $t(\mathbf{X})$, we adapt the method
recommended in \cite{1683770} which picks a linear function of the confidence
vector by maximizing the multi-label accuracy on the training set.

\section{Experimental Results}

In this section, we investigate the performance of the CNNs trained for
the disease-targeted and lesion-targeted skin disease classifications,
respectively. For both the disease-targeted and lesion-targeted classifications,
we use transfer learning \cite{DBLP:journals/corr/YosinskiCBL14} \footnote{We
use transfer learning and fine-tuning interchangeably in this paper.} to train
the CNNs. However, note that the ImageNet pre-trained models are trained
from images containing mostly artifacts, animals, and plants. This is very
different from our skin disease cases. To investigate the features learned only
from skin diseases and avoid using useless features, we also train the CNNs
from scratch.

We conduct all the experiments using the Caffe deep learning framework \cite{jia2014caffe}
and run the programs with a GeForce GTX 970 GPU. For the hyper-parameters, 
we follow the settings used by the AlexNet, i.e., batch size $= 256$, 
momentum $= 0.9$ and weight decay $= 5.0e^{-4}$. We use $0.01$ and $0.001$ learning
rate for fine-tuning and training from scratch, respectively.

\subsection{Performance of Disease-Targeted Classification}

\begin{table}
  \renewcommand{\arraystretch}{1.3}
  \caption{Accuracies and MAP of the disease-targeted classification}
  \centering
  \begin{tabular}{c|c|c|c}
  \textbf{Learning Type} & \textbf{Top-1 Accuracy} & \textbf{Top-5 Accuracy} & \textbf{MAP}\\
  \hline\hline
  Fine-tuning            &                  27.6\% &                  57.9\% &         0.42\\
  Scratch                &                  21.1\% &                  48.9\% &         0.35\\
  \end{tabular}
  \label{tab: diagnosis accuracies}
\end{table}

\begin{figure}
  \centering
  \includegraphics[scale=0.40]{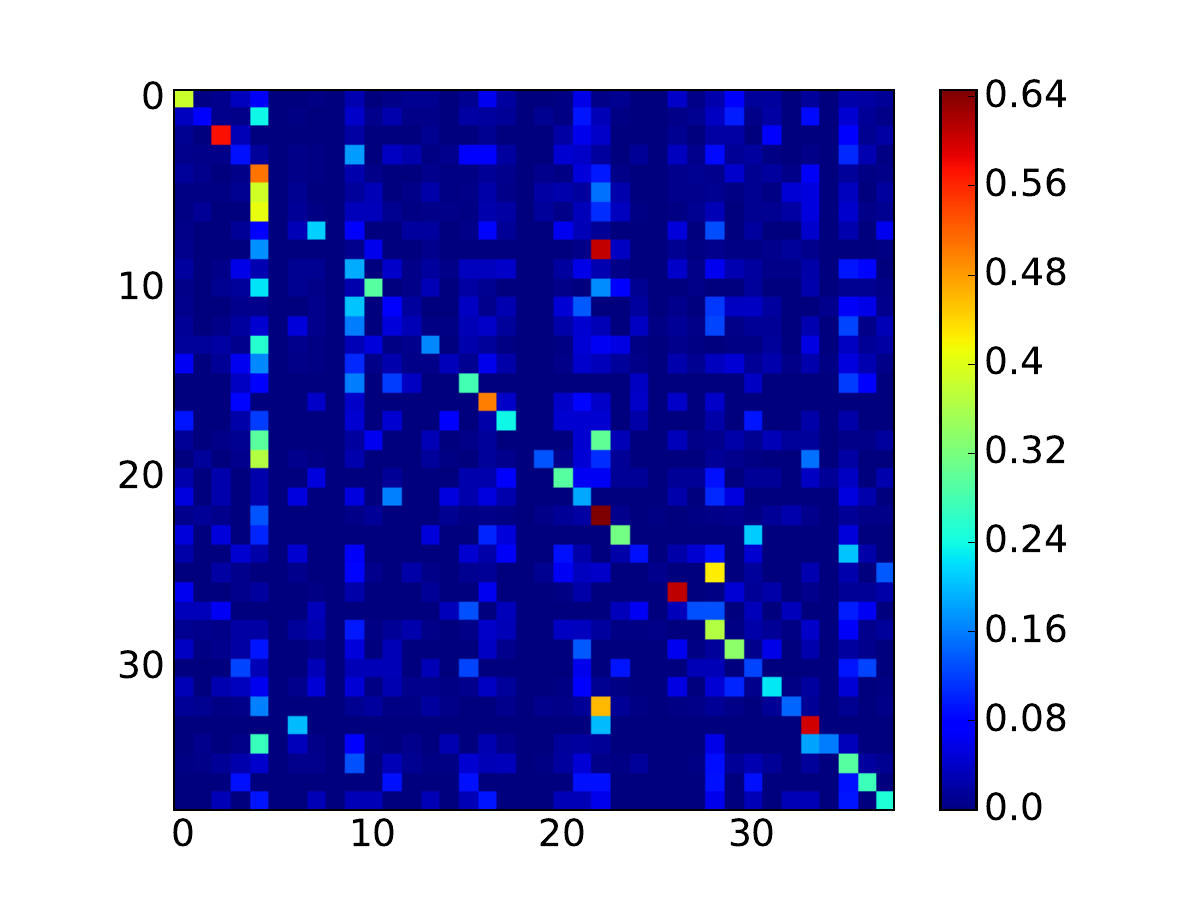}
  \caption{The confusion matrix of the disease-targeted skin disease classifier
  with the CNN trained using fine-tuning. Row: Actual diagnosis. Column: Predicted
  diagnosis.}
  \label{fig: confusion matrix}
\end{figure}

To evaluate the performance of the disease-targeted skin disease classifier,
we use the top-$1$ and top-$5$ accuracies, MAP score, and the confusion matrix as
the metrics. Following the notations in Section \ref{sec: methodology}, let
$\mathbf{C}_n$ be the output of the multi-class CNN when the input is $\mathbf{X}_n$
and $\mathbf{T}_n^k$ be the labels of the $k$ largest elements in $\mathbf{C}_n$.
The top-$k$ accuracy of the multi-class CNN on the test set is given as
\begin{equation}
A_{\text{top-}k} = \frac{\sum_{n=1}^NZ_n^k}{N},
\end{equation}
where $Z_n^k$ is
\begin{equation}
Z_n^k = \begin{cases}
1, & y^n \in \mathbf{T}_n^k, \\
0, & \text{otherwise.}
\end{cases}
\end{equation}
and $N$ is the total number of images in the test set. For the MAP, we adapt the
definition described in \cite{zhu2004recall}:
\begin{equation}
\text{MAP} = \frac{1}{N}\sum_{i=1}^N\sum_{j=1}^Qp_i(j)\Delta r_i(j),
\end{equation}
where $p_i(j)$ and $r_i(j)$ denote the precision and recall of the $i$th image at
fraction $j$, $\Delta r_i(j)$ denotes the change in recall from $j-1$ to $j$ and
$Q$ is the total number of possible lesions.
Finally, for the confusion matrix $\mathbf{M}$, its elements are given as
\begin{equation} \label{eq: confusion matrix}
\mathbf{M}(i, j) = \frac{\sum_{n=1}^NI(y^n = i)I(\widehat{y}^n = j)}{N_i}
\end{equation}
where $y^n$ is the ground truth, $\widehat{y}^n$ is the prediction and $N_i$ is
the number of images whose ground truth is $i$.

Table \ref{tab: diagnosis accuracies} shows the accuracies and MAP
of the disease-targeted skin disease classifiers with the CNNs trained from
scratch or using fine-tuning. It is interesting to note that the CNN trained using transfer
learning performs better than the CNN trained from scratch only on skin diseases. It suggests that the
more general features learned from the richer set of ImageNet images can still benefit the more specific classification
of the skin diseases. And training from scratch did not necessarily help the CNN learn more
useful features related to the skin diseases. However, even for the CNN trained
with fine-tuning, the accuracies and MAP are not satisfactory. Only $27.6\%$ top-$1$
accuracy, $57.9\%$ top-$5$ accuracy, and $0.42$ MAP score are achieved.

The confusion matrix computed for the fine-tuned CNN is given in Figure
\ref{fig: confusion matrix}. The row indices correspond to the actual diagnosis
labels and the column indices denote the predicted diagnosis labels. Each cell
is computed using Equation \eqref{eq: confusion matrix} which is the
percentage of the prediction $j$ among images with ground truth $i$. A good
multi-class classifier should have high diagonal values. We find in Figure
\ref{fig: confusion matrix} that there are some off-diagonal cells with relatively
high values. This is because some skin diseases are visually similar, and
the CNNs trained with diagnosis labels still cannot distinguish among them. For example,
the off-diagonal cell at row $8$ and column $22$ has a value of $0.60$. Here,
label $8$ represents ``compound nevus'' and label $22$ stands for
``malignant melanoma''. It means about $60\%$ of the ``compound nevus'' images are
incorrectly labeled as ``malignant melanoma''. If we look at the two images
in the first column of Figure \ref{fig: similar diseases}, we can see that these
two diseases look so similar in appearance that not surprisingly the disease-targeted classifier
fails to distinguish them.

\begin{figure}
  \centering
  \includegraphics[scale=0.24]{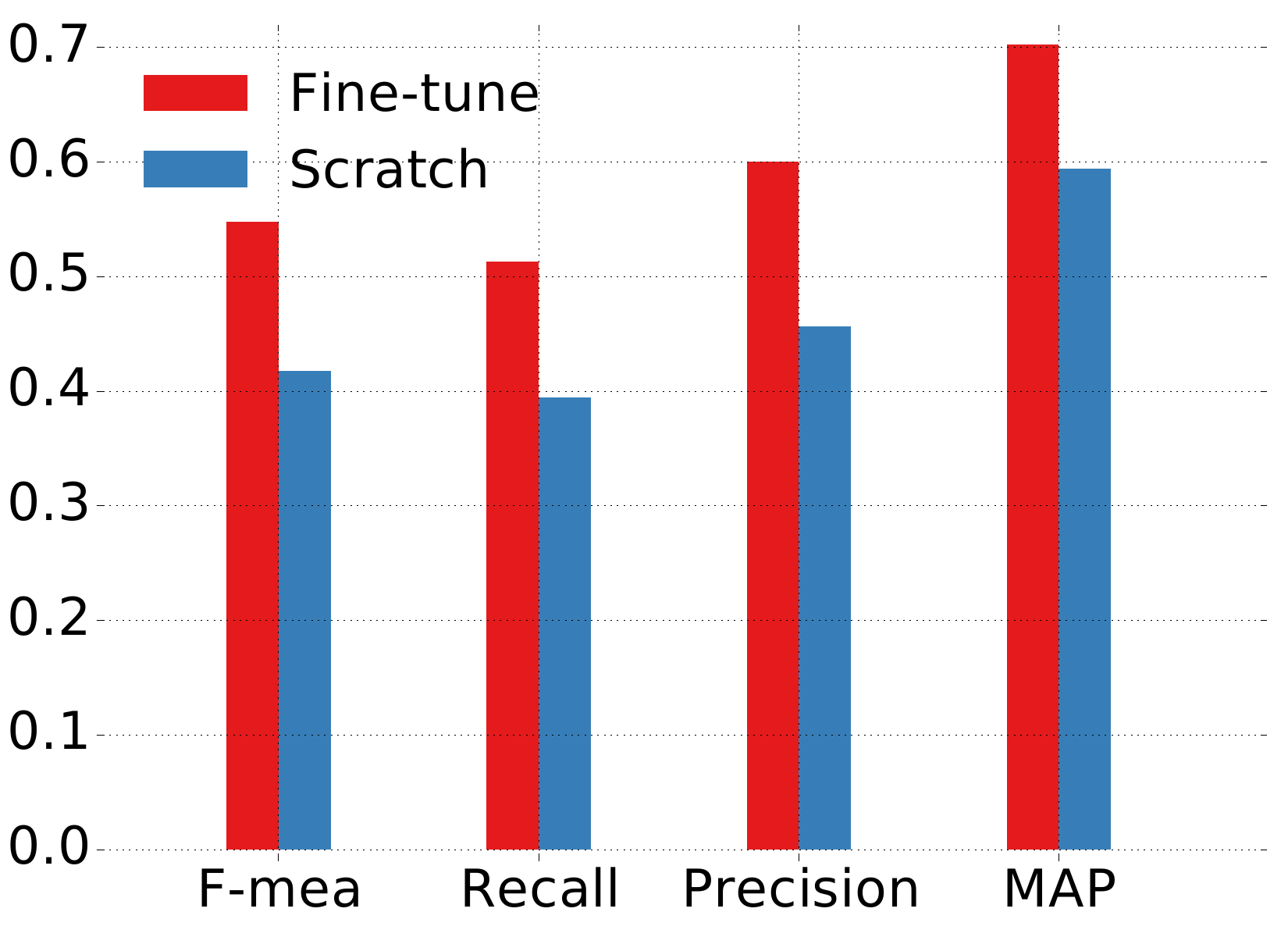}
  \caption{Macro-average of precision, recalls, and F-measures as well as MAP.}
  \label{fig: overall metrics}
\end{figure}

\subsection{Performance of Lesion-Targeted Classification}

\begin{figure*}
  \centering
  \subfloat[Precisions]
    {\includegraphics[width=0.32\linewidth]{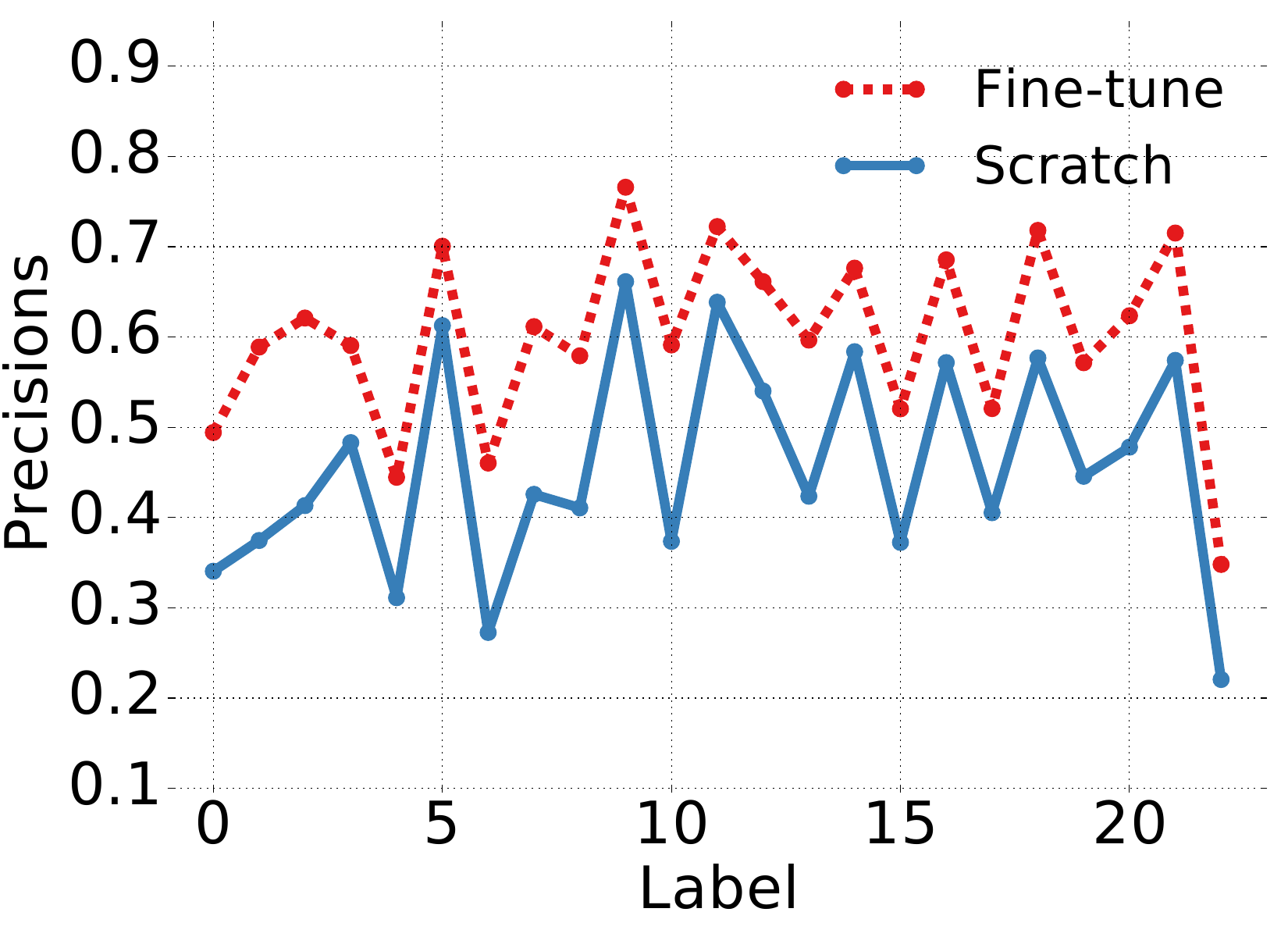}}\hfill
  \subfloat[Recalls]
    {\includegraphics[width=0.32\linewidth]{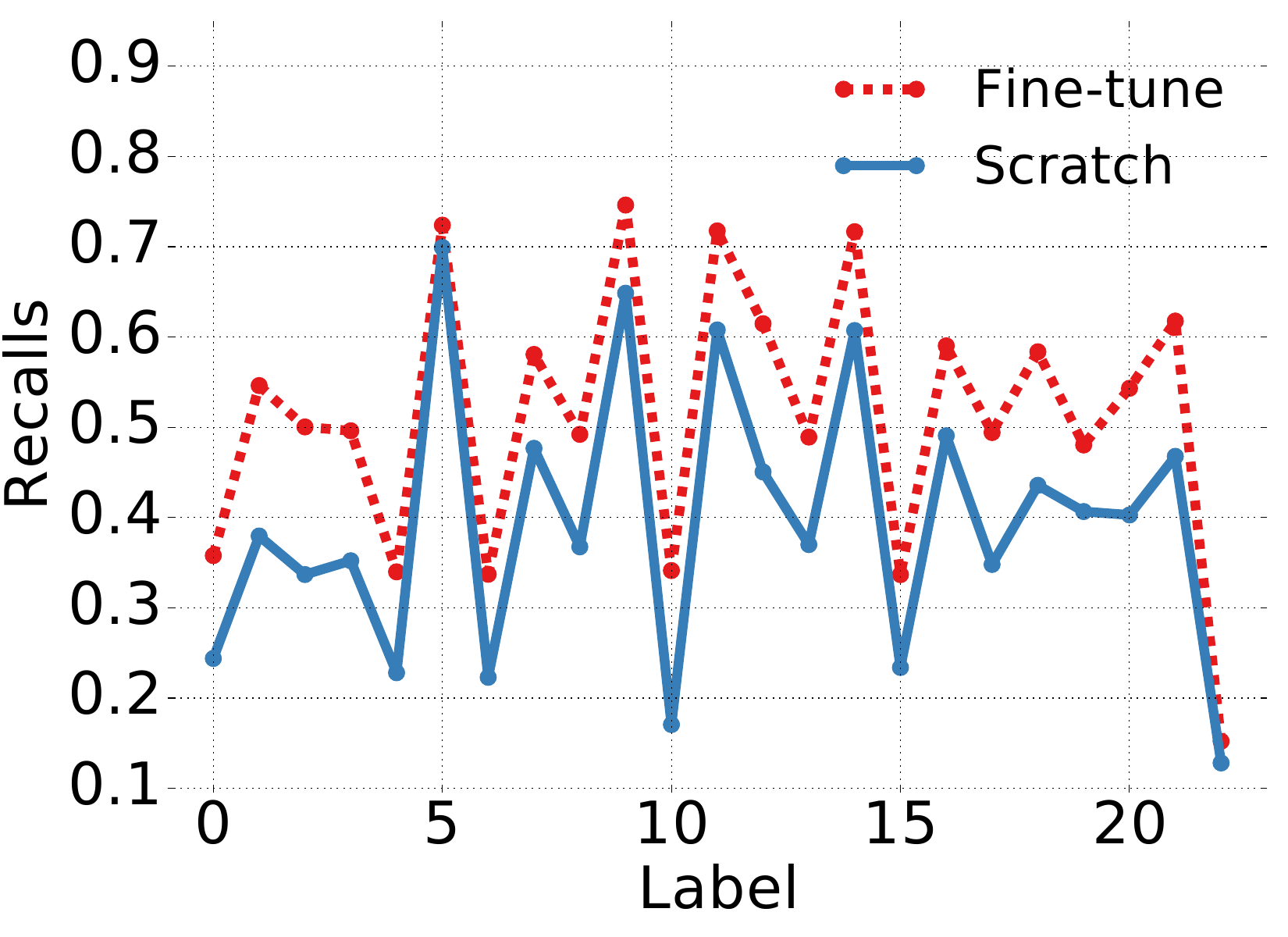}}\hfill
  \subfloat[F-measures]
    {\includegraphics[width=0.32\linewidth]{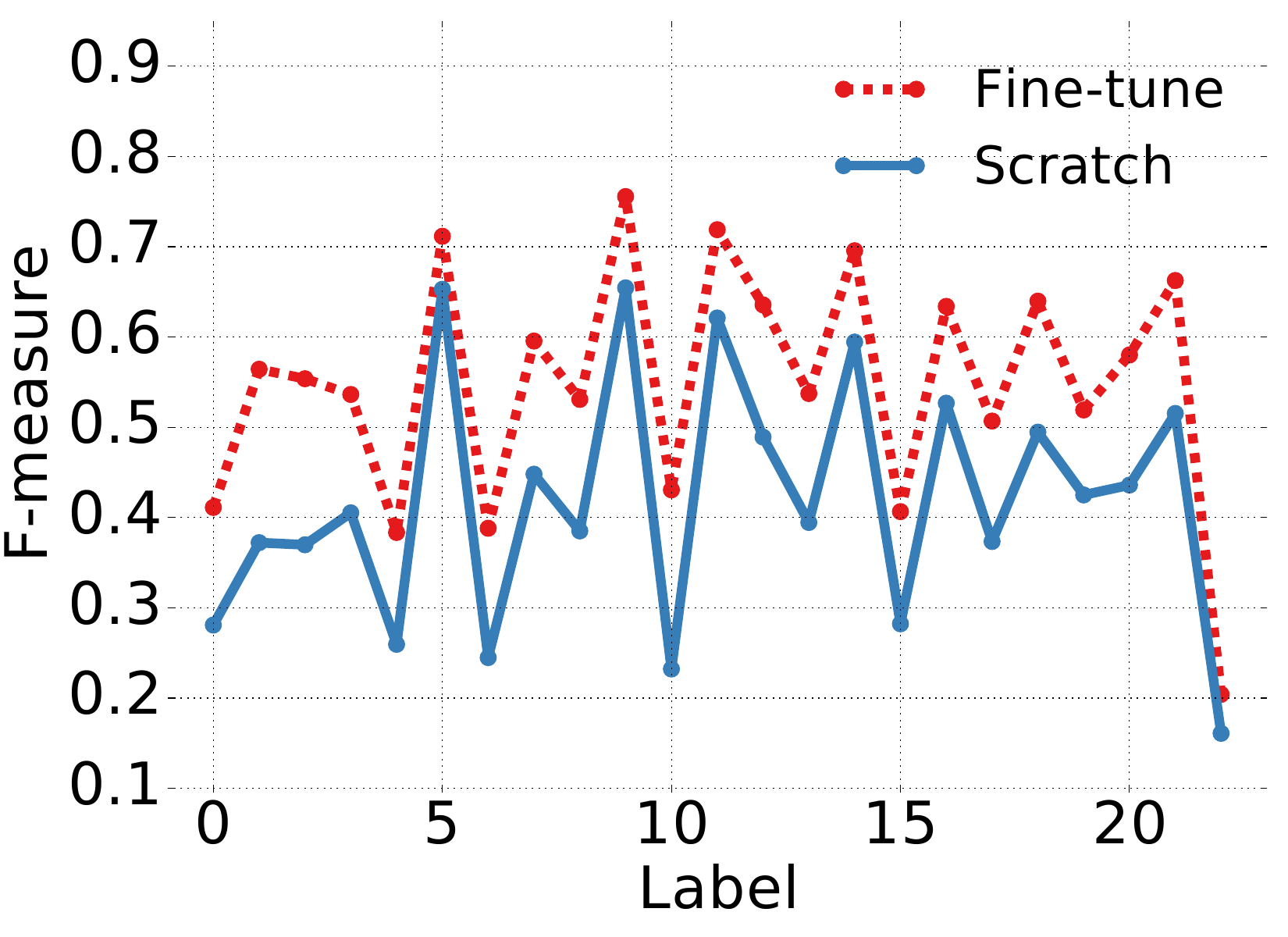}}\hfill
  \caption{Label-based precisions, recalls, and f-measures}
  \label{fig: precision, recall, f-mea}
\end{figure*}

As we use a multi-label classifier for the lesion-targeted skin disease classification,
the evaluation metrics used in this experiment are different from those used
in the previous section. To evaluate the performance of the classifier on each
label, we use the label-based precision, recall and F-measure. And to evaluate 
the overall performance, we use the macro-average of the precision, recall and F-measure.
In addition, the MAP is also used as an evaluation metric of the overall performance.

Let $Y_i$ be the set of images whose ground truth contains lesion $i$ and $Z_i$ be
the set of images whose prediction contains lesion $i$. Then, the label-based
and the macro-averaged precision, recall, and F-measure can be defined as
\begin{align}
\begin{split}
P_i &= \frac{|Y_i \cap Z_i|}{|Z_i|}, P_{macro} = \frac{1}{Q}\sum_{i=1}^QP_i, \\
R_i &= \frac{|Y_i \cap Z_i|}{|Y_i|}, R_{macro} = \frac{1}{Q}\sum_{i=1}^QR_i, \\
F_i &= \frac{2|Y_i||Z_i|}{|Y_i| + |Z_i|}, F_{macro} = \frac{1}{Q}\sum_{i=1}^QF_i. \\
\end{split}
\end{align}
where $Q$ is the total number of possible lesion tags.

Figure \ref{fig: overall metrics} shows the overall performance of the lesion-targeted
skin disease classifiers. The macro-average of the F-measure is around $0.55$
and the mean average precision is about $0.70$. This is quite good for
a multi-label problem. The label-based precisions, recalls, and F-measures are
given in Figure \ref{fig: precision, recall, f-mea}. We can see that for the
lesion-targeted skin disease classification, the fine-tuned CNN performs better than the
CNN trained from scratch which is consistent with our observation in Table
\ref{tab: diagnosis accuracies}. It means for the lesion-targeted skin disease classification
problem, it is still beneficial to initialize with weights from ImageNet pretrained
models. We also see that the label-based metrics are mostly above $0.5$ in
the fine-tuning case. Some exceptions are atrophy ($0$), erythemato-squamous ($4$),
excoriation ($6$), oozing ($15$), and vesicle ($22$). The failures are mostly
due to
\begin{inparaenum}[1)]
  \item the lesiona not visually salient or masked by other larger lesions, or
  \item sloppy labeling of the ground truth.
\end{inparaenum}

\begin{figure}
  \centering
  \includegraphics[scale=0.2]{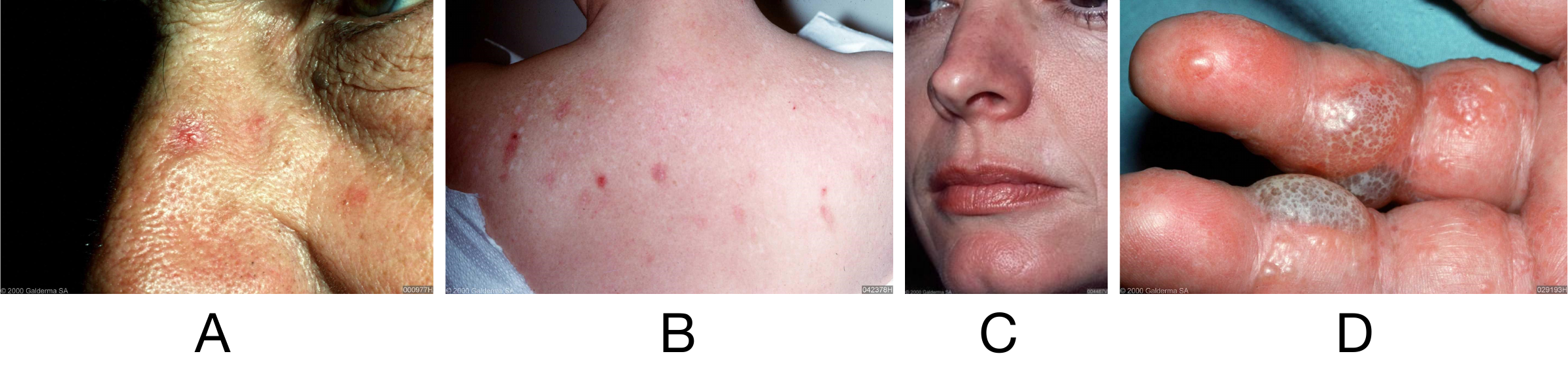}
  \caption{Failure cases. Ground truth (left to right): atrophy, excoriation,
  hypopigmented, vesicle. Top prediction (left to right): erythematous,
  erythematous, ulceration, edema.}
   \vskip -0.1in
  \label{fig: failures}
\end{figure}

Some failure cases are shown in Figure \ref{fig: failures}. Image $A$ is labeled
as atrophy. However, the atrophic characteristic is not so obvious and it is more
like an erythematous lesion. For image $B$, the ground truth is excoriation which
is the little white scars on the back. However, the red erythematous lesion
is more apparent. So the CNN incorrectly classified it as a erythematous lesion.
Similar case can be found in image $D$. For image $C$, the ground truth is actually
incorrect.

\begin{figure}[t]
  \centering
  \includegraphics[scale=0.04]{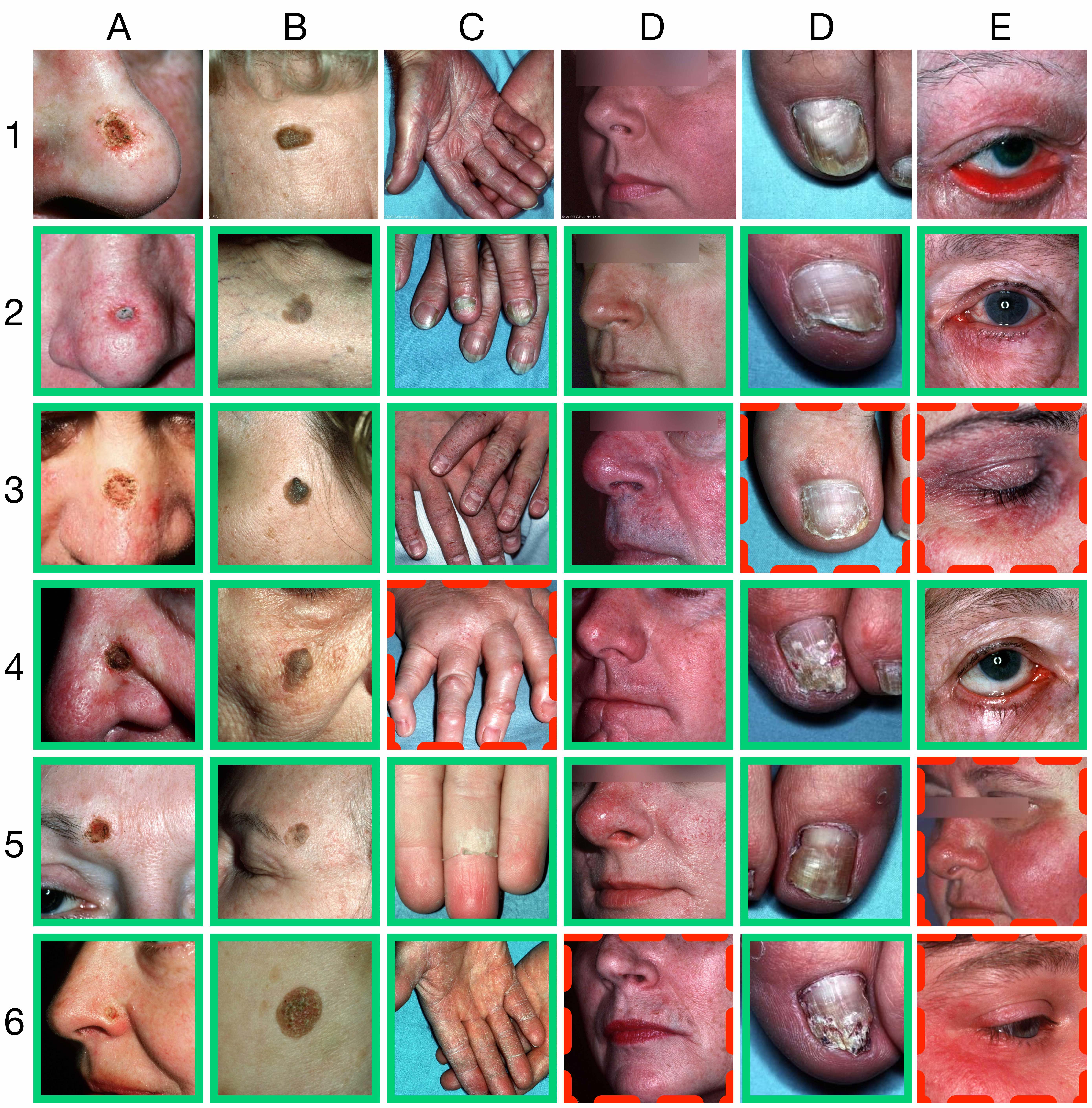}
  \caption{Images retrieved by the lesion-targeted classifier. Row 1: the query
  images from the test set. Row 2-6: the retrieved images from the training set. Dotted borders annotate errors.
  Ground truth of the test images from column A to D: (crust, ulceration),
  (hyperpigmented, tumour), (scales), (erythematous, telangiectasis),
  (nail hyperpigmentation, onycholysis), (edema, erythematous).}
  \label{fig: image retrievals}
  \vskip -0.1in
\end{figure}

Figure \ref{fig: image retrievals} shows the image retrievals using the lesion-targeted
classifier. Here, we take the output of the second to last fully-connected layer
($4096$ dimension) as the feature vector. For each query image from the test set,
we compare its features with all the images in the training set and outputs the
$5$-nearest neighbors (in euclidean distance) as the retrievals. The retrieved
images with green solid frames match at least one lesion tag of the query
image. And those images with red dashed frames have no common lesion tags with
the query image. We can see that the retrieved images are visually and semantically
similar to the query images.

\section{Conclusion}

In this study, we have showed that, for skin disease
classification using CNNs, lesion tags rather than the diagnosis tags should be considered as the target for automated analysis. To achieve better
diagnosis results, computer aided skin disease diagnosis systems could use
lesion-targeted CNNs as the cornerstone component to facilitate the final disease diagnosis in
conjunction with other evidences. We have built a large-scale dermatology dataset from six
professional photosharing dermatology atlantes. We have trained and tested the
disease-targeted and lesion-targeted classifiers using CNNs. Both fine-tuning
and training from scratch were investigated in training the CNN models. We found that, for
skin disease images, CNNs fine-tuned from pre-trained models perform better
than those trained from scratch. For the disease-targeted classification,
it can only achieve $27.6\%$ top-$1$ accuracy and $57.9\%$ top-$5$ accuracy as
well as $0.42$ MAP. The corresponding confusion matrix contains
some high off-diagonal values which indicates that some skin diseases cannot
be distinguished using diagnosis labels. For the lesion-targeted classification,
a $0.70$ MAP score is achieved, which is remarkable 
for a multi-label classification problem. Image retrieval results also confirm 
that CNNs trained using lesion tags learn the dermatology features very well.

\section*{Acknowledgment}

We gratefully thank the support from the University, and New York
State through Goergen Institute for Data Science.



\bibliographystyle{IEEEtran}
\bibliography{references}


% \end{thebibliography}

\end{document}